\documentclass{article}

\usepackage{microtype}
\usepackage{graphicx}
\usepackage{subfigure}
\usepackage{booktabs} 

\usepackage{hyperref}

\usepackage[accepted]{icml2020}

\usepackage{amsthm,amssymb}
\usepackage{thmtools,thm-restate}

\usepackage{amsmath,amsfonts,bm}

\def\eqref#1{equation~\ref{#1}}

\def\1{\bm{1}}

\def\vb{{\bm{b}}}

\def\vq{{\bm{q}}}

\def\vx{{\bm{x}}}
\def\vy{{\bm{y}}}

\DeclareMathAlphabet{\mathsfit}{\encodingdefault}{\sfdefault}{m}{sl}
\SetMathAlphabet{\mathsfit}{bold}{\encodingdefault}{\sfdefault}{bx}{n}

\def\gB{{\mathcal{B}}}

\def\gD{{\mathcal{D}}}

\def\gF{{\mathcal{F}}}

\def\gJ{{\mathcal{J}}}

\def\gL{{\mathcal{L}}}

\def\gN{{\mathcal{N}}}

\newcommand{\E}{\mathbb{E}}

\newcommand{\R}{\mathbb{R}}

\DeclareMathOperator*{\argmax}{arg\,max}
\DeclareMathOperator*{\argmin}{arg\,min}

\newcommand{\rbr}[1]{\left(#1\right)}
\newcommand{\sbr}[1]{\left[#1\right]}
\newcommand{\cbr}[1]{\left\{#1\right\}}

\usepackage{graphicx}
\usepackage{color}
\usepackage{cite}
\usepackage{enumitem}
\usepackage{soul}
\usepackage{algorithmic}
\usepackage[algo2e,vlined,linesnumbered,ruled,resetcount]{algorithm2e}

\usepackage[export]{adjustbox}
\newcommand*{\Scale}[2][4]{\scalebox{#1}{$#2$}}%
\usepackage{xr-hyper}
\makeatletter
\newcommand*{\addFileDependency}[1]{
  \typeout{(#1)}
  \@addtofilelist{#1}
  \IfFileExists{#1}{}{\typeout{No file #1.}}
}
\makeatother

\icmltitlerunning{Learning to Stop While Learning to Predict}

\begin{document}

\twocolumn[
\icmltitle{
Learning to Stop While Learning to Predict
}





\begin{icmlauthorlist}
\icmlauthor{Xinshi Chen}{gt}
\icmlauthor{Hanjun Dai}{google}
\icmlauthor{Yu Li}{kaust}
\icmlauthor{Xin Gao}{kaust}
\icmlauthor{Le Song}{gt,af}
\end{icmlauthorlist}

\icmlaffiliation{gt}{Georgia Institute of Technology, USA}
\icmlaffiliation{google}{Google Research, USA}
\icmlaffiliation{kaust}{King Abdullah University of Science and Technology, Saudi Arabia}
\icmlaffiliation{af}{Ant Financial, China}

\icmlcorrespondingauthor{Xinshi Chen}{xinshi.chen@gatech.edu}
\icmlcorrespondingauthor{Le Song}{lsong@cc.gatech.edu}


\icmlkeywords{Machine Learning, ICML}

\vskip 0.3in
]

\printAffiliationsAndNotice{}

\begin{abstract}
There is a recent surge of interest in designing deep architectures based on the update steps in traditional algorithms, or learning neural networks to improve and replace traditional algorithms. While traditional algorithms have certain stopping criteria for outputting results at different iterations, many algorithm-inspired deep models are restricted to a ``fixed-depth'' for all inputs. Similar to algorithms, the optimal depth of a deep architecture may be different for different input instances, either to avoid ``over-thinking'', or because we want to compute less for operations converged already. In this paper, we tackle this varying depth problem using a steerable architecture, where a feed-forward deep model and a variational stopping policy are learned together to sequentially determine the optimal number of layers for each input instance. Training such architecture is very challenging. We provide a variational Bayes perspective and design a novel and effective training procedure which decomposes the task into an oracle model learning stage and an imitation stage. Experimentally, we show that the learned deep model along with the stopping policy improves the performances on a diverse set of tasks, including learning sparse recovery, few-shot meta learning, and computer vision tasks.

\end{abstract}

\section{Introduction}
\label{sec:intro}

Recently, researchers are increasingly interested in the connections between deep learning models and traditional algorithms: deep learning models are viewed as parameterized algorithms that operate on each input instance iteratively, and traditional algorithms are used as templates for designing deep learning architectures. While an important concept in traditional algorithms is the stopping criteria for outputting the result, which can be either a \textit{convergence} condition or an \textit{early stopping} rule, such stopping criteria has been more or less ignored in algorithm-inspired deep learning models. 
A ``fixed-depth" deep model is used to operate on all problem instances (Fig.~\ref{fig:motivation} (a)). Intuitively, for deep learning models, the optimal depth (or the optimal number of steps to operate on an input) can also be different for different input instances, either because  we want to compute less for operations converged already, or we want to generalize better by avoiding ``over-thinking''. Such motivation aligns well with both the cognitive science literature~\citep{jones2009optimal} and many examples below:
\begin{figure}[t!]
    \centering
    \includegraphics[width=\linewidth]{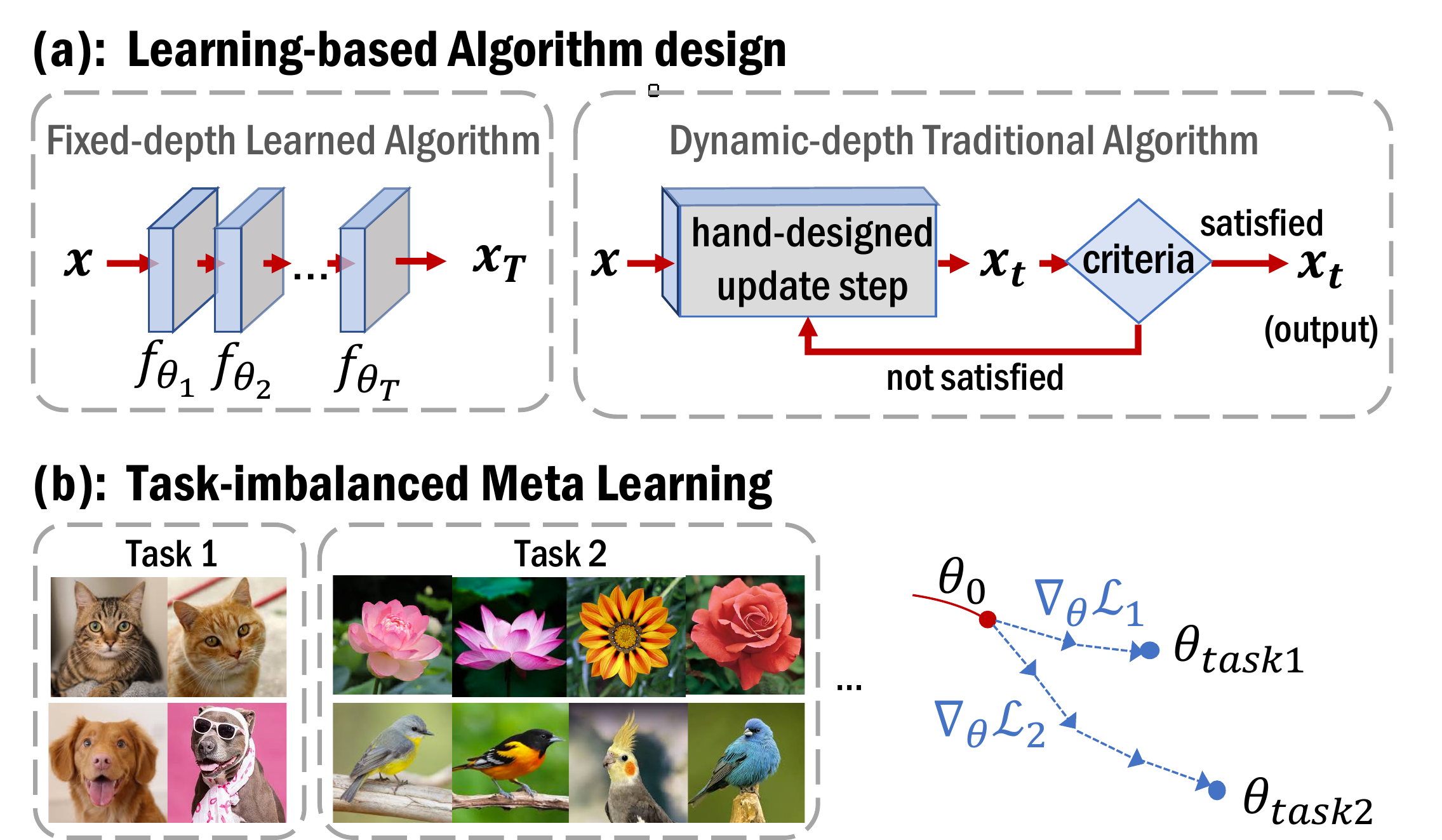}D
    \vspace{-5.5mm}
    \caption{Motivation for learning to stop.}
    \label{fig:motivation}
    \vspace{-3mm}
\end{figure}
\begin{itemize}[leftmargin=*,nolistsep,nosep]
    \item 
    In learning to optimize~\citep{andrychowicz2016learning,li2016learning}, neural networks are used as the optimizer to minimize some loss function. Depending on the initialization and the objective function, an optimizer should converge in different number of steps;
    \item In learning to solve statistical inverse problems such as compressed sensing~\citep{chen2018theoretical,liu2018alista}, inverse covariance estimation~\citep{shrivastava2020glad}, and image denoising~\citep{zhang2018dynamically}, deep models are learned to directly predict the recovery results. In traditional algorithms, problem-dependent early stopping rules are widely used to achieve regularization for a variance-bias trade-off. Deep learning models for solving such problems maybe also achieve a better recovery accuracy by allowing instance-specific computation steps;
    \item In meta learning, MAML~\citep{finn2017model} used an unrolled and parametrized algorithm to adapt a common parameter to a new task. However, depending on the \textit{similarity} of the new task to the old tasks, or, in a more realistic \textit{task-imbalanced} setting where different tasks have different numbers of data points (Fig.~\ref{fig:motivation} (b)), a task-specific number of adaptation steps is more favorable to avoid under or over adaption.  
\end{itemize}

\begin{figure}[t!]
    \centering
    \includegraphics[width=0.96\linewidth]{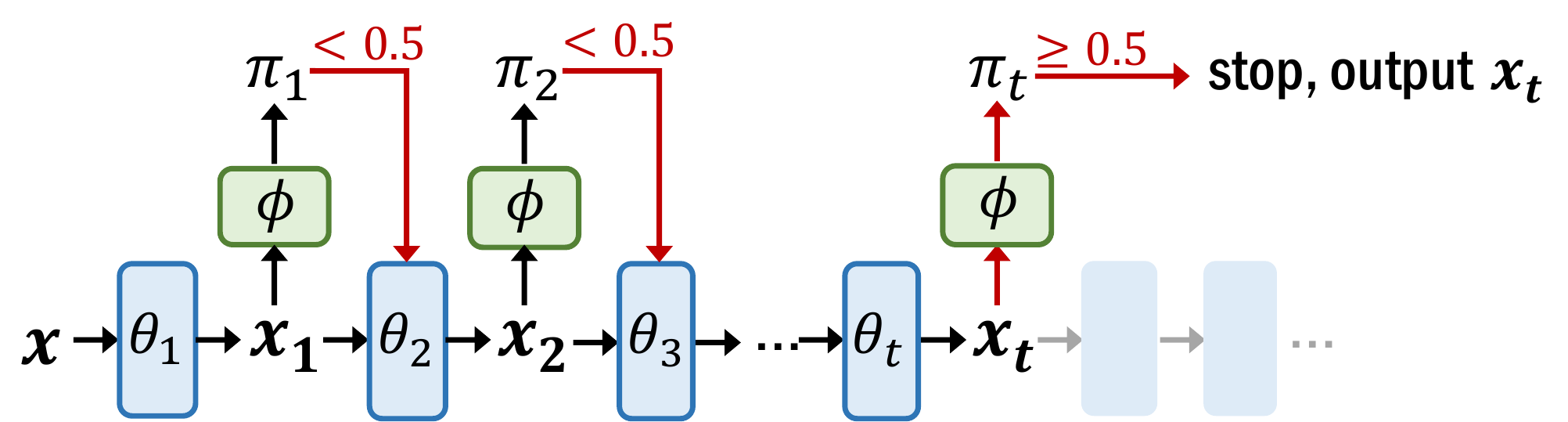}
    \vspace{-3mm}
    \caption{Two-component model: learning to predict (\textit{blue}) while learning to stopping (\textit{green}).}
    \label{fig:model}
    \vspace{-4.5mm}
\end{figure}

To address the varying depth problem, we propose to learn a steerable architecture, where a shared feed-forward model for normal prediction and an additional stopping policy are learned together to sequentially determine the optimal number of layers for each input instance. In our framework, the model consists of (see Fig.~\ref{fig:model})
\begin{itemize}[leftmargin=*,nolistsep,nosep]
    \item A \textbf{feed-forward or recurrent mapping} $\gF_{\theta}$, which transforms the input $\vx$ to generate a path of features (or states) $\vx_1,\cdots,\vx_T$; and
    \item A \textbf{stopping policy} $ \pi_{\phi}:(\vx,\vx_t)\mapsto \pi_t \in[0,1]$, which~sequentially observes the states and then determines the probability of stopping the computation of $\gF_{\theta}$ at layer $t$.
\end{itemize}
These two components allow us to sequentially \textbf{predict} the next targeted state while at the same time determining when to \textbf{stop}. 
In this paper, we propose a single objective function for learning both $\theta$ and $\phi$, and we interpret it from the perspective of variational Bayes, where the stopping time $t$ is viewed as a latent variable conditioned on the input $\vx$. With this interpretation, learning $\theta$ corresponds to maximizing the marginal likelihood, and learning $\phi$ corresponds to the inference step for the latent variable, where a variational distribution $q_{\phi}(t)$ is optimized to approximate the posterior. A natural algorithm for solving this problem could be the Expectation-Maximization (EM) algorithm, which can be very hard to train and inefficient.

\begin{figure}[h!]
    \centering
    \includegraphics[width=0.9\linewidth]{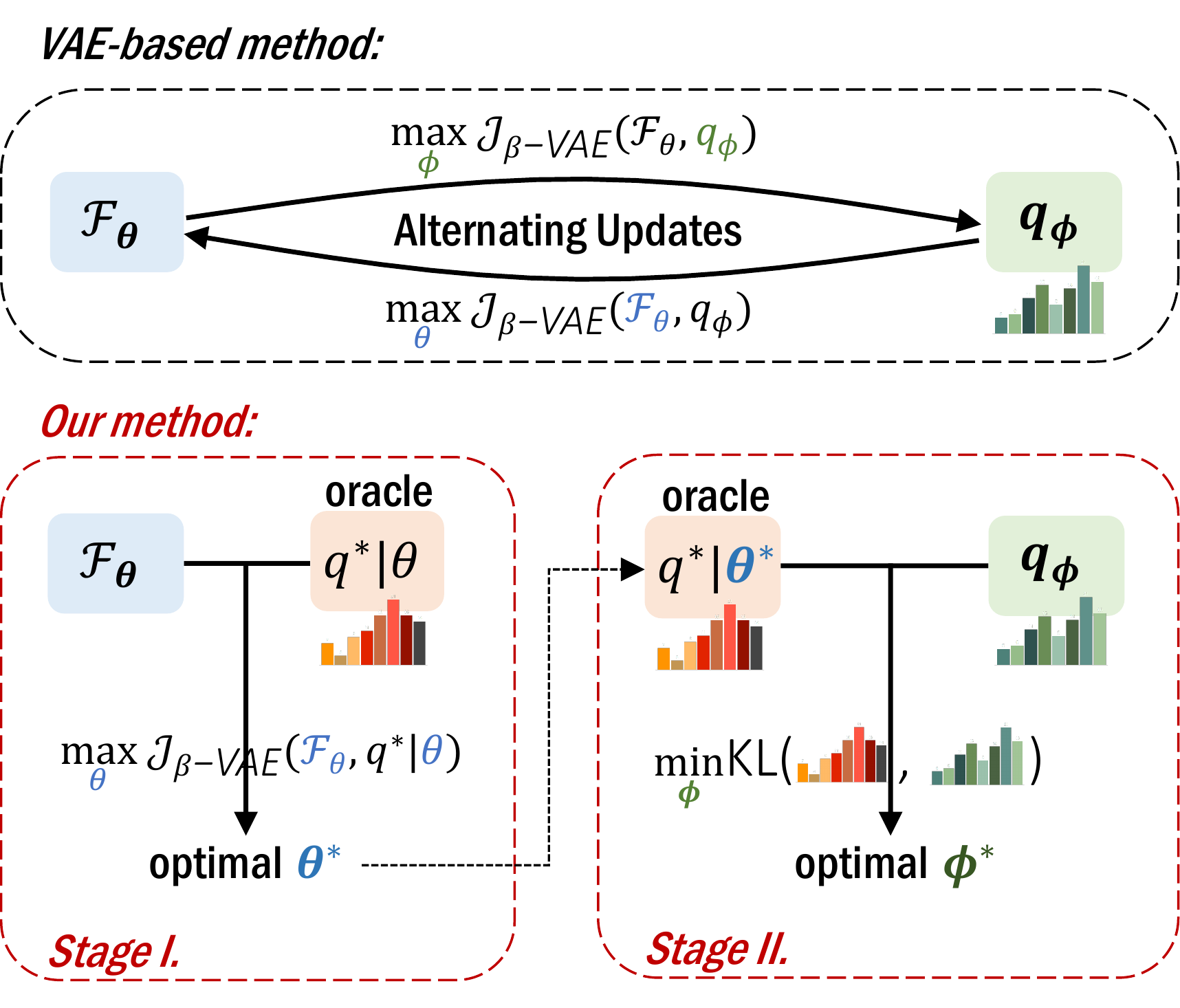}
    \vspace{-3mm}
    \caption{Two-stage training framework.}
    \label{fig:train-frame}
    \vspace{-4.5mm}
\end{figure}

How to learn $\theta$ and $\phi$ effectively and efficiently? We propose a principled and effective training procedure,
where we decompose the task into an oracle model learning stage and an imitation learning stage (Fig.~\ref{fig:train-frame}). More specifically, 
\begin{itemize}[leftmargin=*,nolistsep,nosep]
    \item During the oracle model learning stage,  we utilize a closed-form oracle stopping distribution $q^*|\theta$ which can leverage label information not available at testing time. 
    \item In the imitation learning stage, we use a sequential policy $\pi_\phi$ to mimic the behavior of the oracle policy obtained in the first stage. The sequential policy does not have access to the label so that it can be used during testing phase.
\end{itemize}

This procedure provides us a very good initial predictive model and a stopping policy. We can either directly use these learned models, or  plug them back to the variational EM framework and reiterate to further optimize both together. 

Our proposed learning to stop method is a generic framework that can be applied to a diverse range of applications.
To summarize, our contribution in this paper includes:
\begin{enumerate}[nosep,leftmargin=*]
    \item a variational Bayes perspective to understand the proposed model for learning both the predictive model and the stopping policy together;
    \item a principled and efficient algorithm for jointly learning the predictive model and the stopping policy; 
    and the relation of this algorithm to reinforcement learning;
    \item promising experiments on various tasks including learning to solve sparse recovery problems, task-imbalanced few-shot meta learning, and computer vision tasks,  where we demonstrate the effectiveness of our method in terms of both the prediction accuracy and inference efficiency.
\end{enumerate}

\section{Related Works}

\textbf{Unrolled algorithm.} 
A line of recent works unfold and truncate iterative algorithms to design neural architectures. These algorithm-based deep models can be used to automatically learn a better algorithm from data. This idea has been demonstrated in different problems including sparse signal recovery \citep{gregor2010learning,sun2016deep,borgerding2017amp,metzler2017learned,zhang2018ista,chen2018theoretical,liu2018alista}, sparse inverse covariance estimation~\citep{shrivastava2020glad}, sequential Bayesian inference \citep{chen2019particle}, parameter learning in graphical models \citep{domke2011parameter}, non-negative matrix factorization \citep{yakar2013bilevel}, etc. Unrolled algorithm based deep module has also be used for structured prediction \citep{belanger2017end,ingraham2018learning,chen2020rna}.  Before the training phase, all these works need to assign a fixed number of iterations that is used for every input instance regardless of their varying difficulty level. Our proposed method is orthogonal and complementary to all these works, by taking the variety of the input instances into account via  adaptive stopping time.

\textbf{Meta learning.} Optimization-based meta learning techniq- ues are widely applied for solving challenging few-shot learning problems \citep{ravi2017optimization,finn2017model,li2017meta}. 
Several recent advances proposed task-adaptive meta-learning models which incorporate  task-specific parameters \citep{qiao2018few,lee2018gradient,Na2020Learning} or task-dependent metric scaling \citep{oreshkin2018tadam}. 
In parallel with these task-adaptive methods, we propose a task-specific number of adaptation steps and demonstrate the effectiveness of this simple modification under the task-imbalanced scenarios.

\textbf{Other adaptive-depth deep models.} In image recognition,  `early exits' is proposed mainly aimed at  improving the computation efficiency during the inference phase \citep{teerapittayanon2016branchynet,zamir2017feedback,huang2018multiscale}, but these methods are based on specific architectures.
\citet{kaya2019shallow} proposed to avoiding ``over-thinking'' by early stopping. However, the same as all the other `early exits' models, some heuristic policies are adopted to choose the output layer by confidence scores of internal classifiers. Also, their algorithms for training the feed-forward model $\gF_\theta$ do not take into account the effect of the stopping policy.

\textbf{Optimal stopping.} In optimal control literature, optimal stopping is a problem of choosing a time to take a given action based on sequentially observed random variables in order to maximize
an expected payoff~\citep{shiryaev2007optimal}. When a policy for controlling the evolution of random variables (corresponds to the output of $\gF_{\theta}$) is also involved, it is called a ``mixed control'' problem, which is highly related to our work. Existing works in this area find the optimal controls by solving the Hamilton-Jacobi-Bellman (HJB) equation, which is theoretically grounded \citep{pham1998optimal,ceci2004mixed,dumitrescu2018approximation}. However, they  focus on stochastic differential equation based model and the proposed algorithms suffer from the curse of dimensionality problem. \citet{becker2019deep} use DL to learn the optimal stopping policy, but the learning of $\theta$ is not considered. Besides, \citet{becker2019deep} use  reinforcement learning (RL) to solve the problem. 
In Section~\ref{sec:algo}, we will discuss how our variational inference formulation is related to RL.

\section{Problem Formulation}

In this section, we will introduce how we model the stopping policy together with the predictive deep model, define the joint optimization objective, and interpret this framework from a variational Bayes perspective.

\subsection{Steerable Model}

The predictive model, $\gF_{\theta}$, is a typical $T$-layer deep~model that generates a path of embeddings $(\vx_1,\cdots,\vx_T)$ through:
\begin{align}
  \textbf{Predictive model:} \quad \vx_{t} = f_{\theta_t}(\vx_{t-1}),\, \text{for }t \Scale[0.8]{=1,\cdots,T} \label{eq:dnn-op}
\end{align}
where the initial $\vx_0$ is determined by the input $\vx$.
We denote it by $\gF_{\theta}=\cbr{f_{\theta_1},\cdots,f_{\theta_T}}$ where $\theta\in\Theta$ are the parameters. 
Standard supervised learning methods learn $\theta$ by optimizing an objective estimated on the final state $\vx_T$.
In our~model,
the operations in Eq.~\ref{eq:dnn-op} can be stopped earlier, 
and for different input instance $\vx$, the stopping time $t$ can be different.

Our stopping policy, $\pi_\phi$, determines whether to stop at $t$-th step after observing the input $\vx$ and its first $t$ states $\vx_{1:t}$ transformed by $\gF_\theta$. If we assume the Markov property, then $\pi_\phi$ only needs to observe the most recent state $\vx_t$.
In this paper, we only input $\vx$ and $\vx_t$ to $\pi_\phi$ at each step $t$, but it is trivial to generalize it $\pi_\pi(\vx,\vx_{1:t})$. More precisely,
 $\pi_\phi$ is defined as a randomized policy as follows:
\begin{align}
  \textbf{Stopping policy:}~~ \pi_t = \pi_{\phi}(\vx,\vx_t),\, \text{for } t\Scale[0.8]{=1,\cdots,T-1} \label{eq:def-stop}
\end{align}
where $\pi_t\in[0,1]$ is the probability of stopping. We abuse the notation $\pi$ to both represent the parametrized policy and also the probability mass. 

This stopping policy sequentially makes a decision whenever a new state $\vx_t$ is observed. Conditioned on the states observed until step $t$, whether to stop before $t$ is independent on states after $t$. Therefore, once it decides to stop at $t$, the remaining computations can be saved, which is a favorable property when the inference time is a concern, or for some optimal stopping problems such as option trading where getting back to earlier states is not allowed.

\subsection{From Sequential Policy To Stop Time Distribution}

The stopping policy $\pi_\phi$ makes sequential actions based on the observations, where $\pi_t:=\pi_\phi(\vx,\vx_t)$ is the probability of stopping when $\vx_t$ is observed. These sequential actions $\pi_1,\cdots,\pi_{T-1}$ jointly determines the random time $t$ at which
the stop occurs. Induced by $\pi_\phi$, the probability mass function of the stop time $t$, denoted as $q_\phi$, can be computed~by 
\begin{align}
\textbf{Variational}&\textbf{ stop time distribution:}\nonumber \\
&\begin{cases}
q_{\phi}(t)=
    \pi_t\prod_{\tau=1}^{t-1}(1-\pi_\tau)&\text{if}~ t<T, \\
q_\phi(T)=\prod_{\tau=1}^{T-1}(1-\pi_\tau) &\text{else}.
\end{cases}\label{eq:q_dist}
\end{align}
In this equation, the product $\prod_{\tau=1}^{t-1}(1-\pi_\tau)$ indicates the probability of `not stopped before $t$', which is the survival probability. Multiply this survival probability with $\pi_t$, we have the stop time distribution $q_\phi(t)$. For the last time step $T$,
the stop probability $q_\phi(T)$ simply equals to the survival probability at $T$, which means if the process is `not stopped before $T$', then it must stop at $T$.

Note that we only use $\pi_\phi$ in our model to sequentially determine whether to stop. However, we use the induced probability mass $q_\phi$ to help design the training objective and also the algorithm.

\subsection{Optimization Objective}

Note that the stop time $t$ is a discrete random variable with distribution determined by $q_{\phi}(t)$. Given the observed label $\vy$ of an input $\vx$, the loss of the predictive model stopped~at position $t$ can computed as  $\ell(\vy,\vx_t;\theta)$ where $\ell(\cdot)$ is a loss function. Taking into account all possible stopping positions, we will be interested in the loss in expectation over $t$,
\begin{align}
   \gL(\theta,q_\phi;\vx,\vy):=\E_{t\sim q_{\phi}}\ell(\vy, \vx_t ; \theta) - {\beta} H(q_{\phi}),    \label{eq:joint-loss}
\end{align}
where $H(q_{\phi}):= -\sum_t q_{\phi}(t) \log q_{\phi}(t)$ is an entropy regularization and $\beta$ is the regularization coefficient. 
Given a data set $\gD=\{(\vx,\vy)\}$, the parameters of the predictive model and the stopping policy can be estimated by
\begin{align}
    \textstyle \min_{\theta, \phi}~~ \frac{1}{|\gD|}\sum_{(\vx,\vy)\in\gD} \gL(\theta,q_\phi;\vx,\vy).
\end{align}
To better interpret the model and objective, in the following, we will make a connection from the perspective of variational Bayes, and how the objective function defined in Eq.~\ref{eq:joint-loss} is equivalent to the $\beta$-VAE objective.

\subsection{Variational Bayes Perspective} 

In the Bayes' framework, a probabilistic model typically consists of prior, likelihood function and posterior of the latent variable. We find the correspondence between our model and a probabilistic model as follows (also see Table~\ref{tab:relation})
\begin{itemize}[leftmargin=*,nosep]
    \item we view the adaptive stopping time $t$ as a \textit{latent variable} which is unobserved;
    \item The conditional \textit{prior}  $p(t|\vx)$ of $t$ is a uniform distribution over all the layers in this paper. However, if one wants to reduce the computation cost and penalize the stopping decisions at deeper layers, a prior with smaller probability on deeper layers can be defined to regularize the results;
    \item The \textit{likelihood} function $p_{\theta}(\vy|t,\vx)$ of the observed label $\vy$ is controlled by $\theta$, since $\gF_\theta$ determines the states $\vx_t$;
    \item The \textit{posterior} distribution over the stopping time $t$ can be computed by Bayes' rule $p_{\theta}(t|\vy,\vx)\propto p_{\theta}(\vy|t,\vx)p(t|\vx)$, but it requires the observation of the label $\vy$, which is infeasible during testing phase.
\end{itemize}
\begin{table}[h!]
    \vspace{-4mm}
    \centering
    \caption{Corresponds between our model and Bayes' model.}
    \begin{tabular}{c|c}
    \toprule
       stop time $t$ & latent variable \\
        label $\vy$ & observation \\
        loss $\ell(\vy,\vx_t;\theta)$ & likelihood $p_\theta(\vy|t,\vx)$ \\
        stop time distribution $q_\phi$  & posterior $p_\theta(t|\vy,\vx)$\\
        regularization &  prior  $p(t|\vx)$ \\
    \bottomrule
    \end{tabular}
    \vspace{-2mm}
    \label{tab:relation}
\end{table}
  In this probabilistic model, we need to learn $\theta$ to better fit the observed data and learn a variational distribution $q_{\phi}$ over $t$ that only takes $\vx$ and the transformed internal states as inputs to approximate the true posterior.

More specifically, the parameters in the likelihood function and the variational posterior can be optimized using the variational autoencoder (VAE) framework
~\citep{kingma2013auto}. Here we consider a generalized version called $\beta$-VAE \citep{higgins2017beta}, and obtain the optimization objective for data point $(\vx,\vy)$
\begin{align}
   \gJ_{\beta\text{-VAE}}(\theta,q_\phi;\vx,\vy):=& \nonumber\\
   \E_{q_{\phi}}\log p_{\theta}(\vy|t,&\vx) - \beta \text{KL}(q_{\phi}(t)||p(t|\vx)), \label{eq:vae} 
\end{align}
where KL$(\cdot ||\cdot)$ is the KL divergence. When $\beta=1$, it becomes the original VAE objective, i.e., the evidence lower bound (ELBO). 
Now we are ready to present the equivalence relation between the $\beta$-VAE objective and the loss defined in Eq.~\ref{eq:joint-loss}. See Appendix~\ref{app:lemma1} for the proof.

\begin{restatable}{lemma}{primelemma}\label{lm:equivalence}
Under assumptions: 
(i) the loss function $\ell$ in Eq.~\ref{eq:joint-loss} is defined as the negative log-likelihood (NLL), i.e.,
\vspace{-1.5mm}
\[
\ell(\vy,\vx_t;\theta):=-\log p_{\theta}(\vy|t,\vx);
\vspace{-1.5mm}
\]
(ii) the prior $p(t|\vx)$ is a uniform distribution over $t$;

then \textbf{minimizing} the loss $\gL$  in Eq.~\ref{eq:joint-loss} is equivalent to \textbf{maximizing} the $\beta$-VAE objective $\gJ_{\beta\text{-VAE}}$ in Eq.~\ref{eq:vae}.
\end{restatable}
For classification problems, the cross-entropy loss is aligned with NLL. For regression problems with mean squared error (MSE) loss, we can define the likelihood as $p_{\theta}(\vy|t,\vx)\sim \gN(\vx_t, I)$. Then the NLL of this Gaussian distribution is 
$    -\log p_{\theta}(\vy|t,\vx) = \frac{1}{2}\|\vy - \vx_t\|_2^2 +C
$, which is equivalent to MSE loss. More generally, we can always define $p_{\theta}(\vy|t,\vx)\propto \exp(-\ell(\vy,\vx_t;\theta))$. 

This VAE view allows us to design a two-step procedure to effectively learn $\theta$ and $\phi$ in the predictive model and stopping policy, which is presented in the next section.

\section{Effective Training Algorithm}
\label{sec:algo}

VAE-based methods perform optimization steps over $\theta$ (M step for learning) and $\phi$ (E step for inference) alternatively until convergence, which has two limitations in our case:
\begin{itemize}[leftmargin=*,nolistsep,nosep]
    \item[i.] The alternating training can be slow to converge and requires tuning the training scheduling; 
    \item[ii.] The inference step for learning $q_{\phi}$ may have the mode collapse problem, which in this case means $q_{\phi}$ only captures the time step $t$ with highest averaged frequency.
\end{itemize}
To overcome these limitations, we design a training procedure followed by an optional fine-tuning stage using the variational lower bound~in Eq.~\ref{eq:vae}. More specifically, 
\begin{itemize}[wide,nosep]
    \item[\underline{Stage I.}] Find the optimal $\theta$ by maximizing the conditional mariginal  likelihood when the stop time distribution follows an oracle distribution $q^*_{\theta}$.
    \item[\underline{Stage II.}] Fix the optimal $\theta$ learned in Stage I, and only learn the distribution $q_{\phi}$ to mimic the oracle by minimizing the KL divergence between $q_{\phi}$ and $q_{\theta}^*$.
    \item[\underline{Stage III.}] (Optional) Fine-tune $\theta$ and $\phi$ jointly towards the joint objective in Eq.~\ref{eq:vae}.
\end{itemize}
The overall algorithm steps are summarized in Algorithm~\ref{algo:overall}.
In the following sections, we will focus on the derivation of the first two training steps. Then we will discuss several methods to further improve the memory and computation efficiency for training. 

\subsection{Oracle Stop Time Distribution}

We first give the definition of the oracle stop time distribution $q_{\theta}^*$. For each fixed $\theta$, we can find a closed-form solution for the optimal $q_{\theta}^*$ that optimizes the joint objective.
\begin{align*}
    \textstyle q_{\theta}^*(\cdot|\vy,\vx) := \argmax_{\vq\in\Delta^{T-1}} \gJ_{\beta\text{-VAE}}(\theta,\vq;\vx,\vy)
\end{align*}
Alternatively,
$q_{\theta}^*(\cdot|\vy,\vx)= \argmin_{q\in\Delta^{T-1}}  \gL(\theta,\vq;\vx,\vy)$.
Under the mild assumptions in Lemma~\ref{lm:equivalence}, these two optimizations lead to the same optimal oracle distribution.
\begin{align}
   \textbf{Oracle stop time}& \textbf{ distribution:} \nonumber \\
    q_{\theta}^*(t|\vy,\vx)&= \frac{p_{\theta}(\vy|t,\vx)^{\frac{1}{\beta}}}{\sum_{t=1}^T p_{\theta}(\vy|t,\vx)^{\frac{1}{\beta}}}\\ &=\frac{\exp(-\frac{1}{\beta}\ell(\vy,\vx_t;\theta))}{\sum_{t=1}^T \exp(-\frac{1}{\beta}\ell(\vy,\vx_t;\theta))}~~~~\quad
\end{align}
This closed-form solution makes it clear that the oracle picks a step $t$ according to the smallest loss or largest likelihood with an exploration coefficient $\beta$. 

\textit{Remark}: When $\beta=1$, $q^*_{\theta}$ is the same as the posterior distribution $p_{\theta}(t|\vy,\vx)\propto p_{\theta}(\vy|t,\vx)p(t|\vx)$.

Note that there are no new parameters in the oracle distribution. Instead, it depends on the parameters $\theta$ in the predictive model. Overall, the oracle $q^*_\theta$ is a function of $\theta$, $t$, $\vy$ and $\vx$ that has a closed-form. Next, we will introduce how we use this oracle in the first two training stages.

\subsection{Stage I. Predictive Model Learning}

In Stage I, we optimize the parameters $\theta$ in the predictive model by taking into account the oracle stop distribution $q^*_\theta$ . This step corresponds to the M step for learning $\theta$, by maximizing the marginal likelihood. The difference with the normal M step is that here $q_{\phi}$ is replaced by the oracle $q^*_{\theta}$ that gives the optimal stopping distribution so that the marginal likelihood is independent on $\phi$. More precisely, stage I finds the optimum of:
\begin{align}
    \max_\theta \frac{1}{|\gD|}\sum_{(\vx,\vy)\in\gD} \sum_{t=1}^T q^*_\theta(t|\vy,\vx)\log p_{\theta}(\vy|t,\vx), \label{eq:obj-step1}
\end{align}
where the summation over $t$ is the expectation of the likelihood, $\E_{t\sim q^*_\theta(t|\vy,\vx)}\log p_{\theta}(\vy|t,\vx)$. Since $q^*_\theta$ has a differentiable closed-form expression in terms of $\theta, \vx, \vy$ and $t$, the gradient can also propagate through $q^*_\theta$, which is also different from the normal M step.

To summarize, in Stage I., we learn the predictive model parameter $\theta$, by assuming that the stop time always follows the best stopping distribution that depends on $\theta$. In this case, the learning of $\theta$ has already taken into account the effect of the data-specific stop time.

However, we note that the oracle $q_\theta^*$ is not in the form of sequential actions as in Eq.~\ref{eq:def-stop} and it requires the access to the true label $\vy$, so it can not be used for testing. However, it plays an important role in obtaining a sequential policy which will be explained next. 

\begin{algorithm}[t]
\caption{Overall Algorithm}\label{algo:overall}
  \DontPrintSemicolon
  \SetKwFunction{Grad}{Grad}
  \SetKwProg{Fn}{Function}{:}{}
  \SetKwFor{uFor}{For}{do}{}
  \SetKwFor{OuFor}{For}{do {\rm$\quad \quad \quad \quad \triangleright$ Optional Step}}{}
  \SetKwFor{SOFor}{For}{do {\rm$\quad \quad \quad \quad \quad \triangleright$ Stage I.}}{}
  \SetKwFor{STFor}{For}{do {\rm$\quad \quad \quad \quad \quad \triangleright$ Stage II.}}{}
  \SetKwFor{ForPar}{For all}{do in parallel}{}
  \SetKwComment{Comment}{$\triangleright$\ }{}
  \SetCommentSty{mycommfont}
  
  Randomly initialized $\theta$ and $\phi$.\;
  
  \SOFor{$itr=1$ to \#iterations}{
      Sample a batch of data points $\gB\sim \gD$. \;
      
      Take an optimization step to update $\theta$ towards the marginal likelihood function defined in Eq.~\ref{eq:obj-step1}. \;
  }
  
  \STFor{$itr=1$ to \#iterations}{
      Sample a batch of data points $\gB\sim \gD$. \;
      
      Take an optimization step to update $\phi$ towards the reverse KL divergence defined in Eq.~\ref{eq:obj-step2}. \;
  }
  
  \OuFor{$itr=1$ to \#iterations}{
      Sample a batch of data points $\gB\sim \gD$. \;
      
      Update both $\theta$ and $\phi$ towards $\beta$-VAE objective in Eq.~\ref{eq:vae}. \;
  }
  \KwRet $\theta$, $\phi$ \;
\end{algorithm}

\subsection{Stage II. Imitation With Sequential Policy}

In Stage II, we learn the sequential policy $\pi_\phi$ that can best mimic the oracle distribution $q^*_\theta$, where $\theta$ is fixed to be the optimal $\theta$ learned in Stage I. The way of doing so is to minimize the divergence between the oracle $q^*_\theta$ and the variational stop time distribution $q_\phi$ induced by $\pi_\phi$ (Eq.~\ref{eq:q_dist}). There are various variational divergence minimization approaches that we can use~\citep{nowozin2016f}. For example, a widely used objective for variational inference is 
the \textbf{reverse KL divergence}:
\begin{align*}
  \textstyle 
  \text{KL}(q_\phi || q^*_\theta) = \sum_{t=1}^T -q_\phi(t)\log q^*_\theta(t|\vy,\vx) -H(q_\phi).
\end{align*}
\textit{Remark.} We write $q_\phi(t)$ instead of $q_\phi(t|\vx_{1:T},\vx)$ for notation simplicity, but $q_\phi$ is dependent on $\vx$ and $\vx_{1:T}$ (Eq.~\ref{eq:q_dist}).

If we rewrite $q_\phi$ using $\pi_1,\cdots,\pi_{T-1}$ as defined in Eq.~\ref{eq:q_dist}, we can find that minimizing the reverse KL is equivalent to finding the optimal policy $\pi_\phi$ in a reinforcement learning (RL) environment, where the state is $\vx_t$, action $a_t\sim\pi_t:=\pi_\phi(\vx,\vx_t)$ is a stop/continue decision, the state transition is determined by $\theta$ and $a_t$, and the reward is defined as
\begin{align*}
    r(\vx_t,a_t;\vy):=\begin{cases} -\beta \ell(\vy,\vx_t;\theta) & \text{if } a_t = 0 \text{ (i.e. stop)}\\
    0 & \text{if }a_t=1 \text{ (i.e. continue)}
    \end{cases}
\end{align*}
where $\ell(\vy,\vx_t;\theta)=-\log p_\theta(\vy|t,\vx)$. More detials and also the derivation are given in Appendix~\ref{sec:kl-rl} to show that minimizing $\text{KL}(q_\phi || q^*_\theta)$ is equivalent to solving the following \textbf{maximum-entropy RL}:
\begin{align*}
 \max_\phi \E_{\pi_\phi}{\textstyle \sum_{t=1}^T} \sbr{r(\vx_t,a_t;\vy)+ H(\pi_t)}.
\end{align*}
In some related literature, optimal stopping problem is often formulated as an RL problem~\citep{becker2019deep}. Above we bridge the connection between our variational inference formulation and the RL-based optimal stopping literature.

Although reverse KL divergence is a widely used objective, it suffers from the mode collapse issue, which in our case may lead to a distribution $q_\phi$ that captures only a common stopping time $t$ for all $\vx$ that on average performs the best, instead of a more spread-out stopping time. Therefore, we consider the \textbf{forward KL divergence}:
\begin{align}
    \text{KL}(q_\phi || q^*_\theta) = -\sum_{t=1}^T  q^*_\theta(t|\vy,\vx) \log q_\phi(t) -H(q^*_\theta),\label{eq:obj-step2}
\end{align}
which is equivalent to the \textbf{cross-entropy} loss, since the term $H(q^*_\theta)$ can be ignored as $\theta$ is fixed in this step. Experimentally, we find forward KL leads to a better performance.

\subsection{The Optional Fine Tuning Stage}

It is easy to see that our two-stage training procedure also has an EM flavor. However, with the oracle $q_\theta^*$ incorporated,
the training of $\theta$ has already taken into account the effect of the optimal stopping distribution. Therefore, we can save a lot of alternation steps.
After the two-stage training, we can fine-tune $\theta$ and $\phi$ jointly towards the $\beta$-VAE objective. 
Experimentally, we find this additional stage does not improve much the performance trained after the first two stages.

\subsection{Implementation Details For Efficient Training}
Since both objectives in oracle learning stage (Eq.~\ref{eq:obj-step1}) and imitation stage (Eq.~\ref{eq:obj-step2}) involve the summation over $T$ layers, the computation and memory costs during training are higher than standard learning methods. The memory issue is especially important in meta learning. In the following, we introduce several ways of improving the training efficiency.

\textbf{Fewer output channels.} Instead of allowing the model to output $\vx_t$ at any layer, we can choose a smaller number of output channels that are evenly placed along with the layers.

\textbf{Stochastic sampling in Step I.} A Monte Carlo method can be used to approximate the expectation over $q^*_\theta$ in Step I. More precisely, for each $(\vx,\vy)$ we can randomly sample a layer $t_s\sim q^*_\theta(t|\vy,\vx)$ from the oracle, and only compute $\log p_\theta(\vy|t_s,\vx)$ at $t_s$, instead of summing over all $t\in [T]$. Note that, in this case, the gradient will not back-propagate through $q^*_\theta(t|\vy,\vx)$.

\textbf{MAP estimate in Step II.} Instead of approximating the distribution $q^*_\theta$, we can approximate the maximum a posterior (MAP) estimate 
$\textstyle \hat{t}(\vx,\vy) = \argmax_{t\in[T]} q^*_\theta (t|\vy,\vx)$
so that the objective for each sample is $-\log q_\theta( \hat{t}(\vx,\vy))$, which does not involve the summation over $t$. Except for efficiency, we also find this MAP estimate can lead to a higher accuracy, by encouraging the learning of $q_\phi$ to focus more on the sample-wise best layer.

\section{Experiments}

We conduct experiments on (i) learning-based algorithm for sparse recovery, (ii) few-shot meta learning, and (iii) image denoising. The comparison is in an ablation study fashion to better examine whether the stopping policy can improve the performances given the same architecture for the predictive model, and whether our training algorithm is more effective compared to the alternating EM algorithm. In the end, we also discuss our exploration of the image recognition task.

\subsection{Learning To Optimize: Sparse Recovery}

We consider a sparse recovery task which aims at recovering $\vx^*\in \R^n$ from its noisy linear measurements $\vb = A\vx^*+\bm{\epsilon}$, where $A\in \R^{m\times n}$, $\bm{\epsilon}\in \R^m$ is Gaussian white noise, and $m\ll n$. A popular approach is to model the problem as the LASSO formulation
$\min_\vx \frac{1}{2}\|\vb-A\vx\|_2^2 + \rho \|\vx\|_1$ and solves it using iterative methods such as the ISTA \citep{blumensath2008iterative} and FISTA \citep{beck2009fast} algorithms. 
We choose the most popular model named Learned ISTA (LISTA) as the baseline and also as our predictive model. LISTA is a $T$-layer network with update steps:\vspace{-1mm}
\begin{align}
    \vx_{t} = \eta_{\lambda_t}(W_t^1\vb + W_t^2 \vx_{t-1}),\quad t=1,\cdots,T,
\end{align} 
where $\theta = \{(\lambda_t,W_t^1,W_t^2)\}_{t=1}^T$ are leanable parameters. 

\textbf{Experiment setting.} We follow 
\citet{chen2018theoretical} to generate the samples. The signal-to-noise ratio (SNR)~for~each sample is uniformly sampled from 20, 30, and 40. 
The~training loss for LISTA is $\sum_{t=1}^T\gamma^{T-t}\|\vx_t-\vx^*\|_2^2$ where $\gamma\leq 1$. It is commonly used for algorithm-based deep learning, so that there is a supervision signal for every layer. For ISTA and FISTA, we use the training set to tune the hyperparameters by grid search. See Appendix~\ref{app:lista} for more details.
\begin{table}[h!]
    \vspace{-4mm}
    \centering
    \caption{Recovery performances of different algorithms/models.}
    \resizebox{.95\linewidth}{!}{
    \begin{tabular}{@{}c@{\hspace{1mm}}|c|c|c|c@{}}
        \toprule
       SNR  & mixed & 20 & 30 & 40   \\
         \midrule
    FISTA $\Scale[0.8]{(T = 100)}$ & -18.96 & -16.75 & -20.46 & -20.97     \\
    ISTA $\Scale[0.8]{(T = 100)}$ & -14.66 & -13.99 & -14.99 & -15.07 \\ 
    \hline
    ISTA $\Scale[0.8]{(T = 20)}$ & -9.17 & -9.12 & -9.24 & -9.16 \\
    FISTA $\Scale[0.8]{(T = 20)}$ & -11.12 & -10.98 & -11.19 & -11.19\\
    LISTA $\Scale[0.8]{(T = 20)}$ & -17.53  & -16.53 & -18.07 & -18.20  \\
    \textbf{LISTA-stop} $\Scale[0.8]{(T \leqslant 20)}$& \textbf{-22.41} & \textbf{-20.29} & \textbf{-23.90} & \textbf{-24.21}  \\
    \bottomrule
    \end{tabular}
    }
    \label{tab:lista_nmse}
    \vspace{-2mm}
\end{table}

\textbf{Recovery performance.} (Table~\ref{tab:lista_nmse}) We report the NMSE (in dB) results for each model/algorithm evaluated on 1000 fixed test samples per SNR level.
    It is revealed in Table~\ref{tab:lista_nmse} that learning-based methods have better recovery performances, especially for the more difficult tasks (i.e. when SNR is 20). Compared to LISTA, our proposed adaptive-stopping method (LISTA-stop) significantly improve recovery performance. Also, LISTA-stop with $\leqslant20$  iterations performs better than ISTA and FISTA with 100 iterations, which indicates a better convergence. 

\begin{figure}[h!]
\vspace{-3mm}
    \centering
    \begin{tabular}{@{}c@{\hspace{3mm}}c@{}}
      \includegraphics[width=0.49\linewidth]{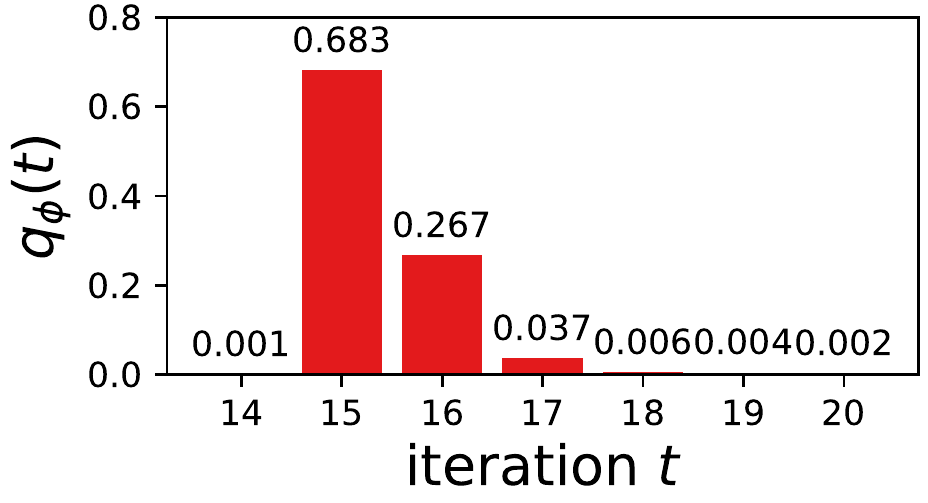}   &
        \includegraphics[width=0.44\linewidth]{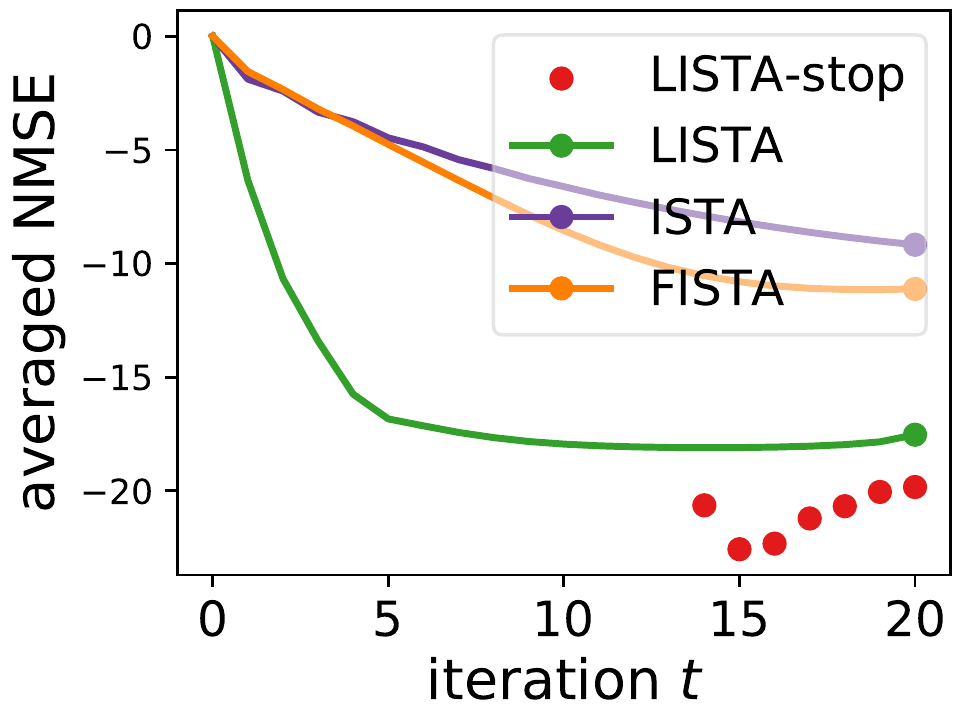}      \vspace{-1mm} \\
       {\small (a) stop time distribution}  & {\small (b) convergence}
    \end{tabular}
    \vspace{-3mm}
    \caption{ \textit{Left}: Stop time distribution $\frac{1}{|\gD_{test}|}\sum_{\vx\in \gD_{test}}q_\phi(t|\vx)$ averaged over the test set. \textit{Right}: Convergence of different algorithms. For LISTA-stop, the NMSE weighted by the stopping distribution $q_\phi$ is plotted. In the first 13 iterations $q_\phi(t)=0$, so no red dots are plotted.}
    \vspace{-3mm}
    \label{fig:q_dist}
\end{figure}

\textbf{Stopping distribution.} The stop time distribution  $q_{\phi}(t)$ induced by  $\pi_\phi$ can be computed via Eq.~\ref{eq:q_dist}. We report in Fig.~\ref{fig:q_dist}  the stopping distribution averaged over the test samples, from which we can see that with a high probability LISTA-stop terminates the process before arriving at 20-th iteration.

\textbf{Convergence comparison.} Fig.~\ref{fig:q_dist} shows the change of NMSE as the number of iterations increases. Since LISTA-stop outputs the results at different iteration steps, it is not meaningful to draw a unified convergence curve. Therefore, we plot the NMSE weighted by the stopping distribution $q_{\phi}$, i.e., $10\log_{10}(\frac{\sum_{i=1}^N q_{\phi}(t|i)
    \|{\vx}_t-\vx^{*,i}\|_2^2}{\sum_{i=1}^N q_{\phi}(t|i)}/ (\frac{\sum_{i=1}^N \|\vx^{*,i}\|_2^2}{N})$, using the red dots. We observe that for LISTA-stop the expected NMSE increases as the number of iterations increase, this might indicate that the later stopped problems are more difficult to solve. Besides, at 15th iteration, the NMSE in Fig.~\ref{fig:q_dist}~(b) is the smallest, while the averaged stop probability mass $q_\phi(15)$ in Fig.~\ref{fig:q_dist}~(a) is the highest.

\begin{table}[h!]
    \vspace{-3mm}
    \centering
    \caption{Different algorithms for training LISTA-stop.}
    \resizebox{.9\linewidth}{!}{
    \begin{tabular}{@{}c|c|c|c|c@{}}
        \toprule
       SNR  & mixed & 20 & 30 & 40   \\
         \midrule
        {AEVB algorithm} & -21.92 & -19.92 & -23.27 & -23.58 \\
    \hline
    Stage I. + II. & \textbf{-22.41} & \textbf{-20.29} & \textbf{-23.90} & \textbf{-24.21}\\
    \hline
    {Stage I.+II.+III.} &  \textbf{-22.78} & \textbf{-20.59} & \textbf{-24.29} & \textbf{-24.73}\\
    \bottomrule
    \end{tabular}}
    \vspace{-1mm}
    \label{tab:ablation-algo}
\end{table}
\textbf{Ablation study on training algorithms.} To show the effectiveness of our two-stage training, in Table~\ref{tab:ablation-algo}, we compare
the results with the auto-encoding variational Bayes (AEVB) algorithm~\citep{anh2018autoencoding} that jointly optimizes $\gF_\theta$ and $q_\phi$. 
We observe that the distribution $q_\phi$ in AEVB gradually~becomes concentrated on one layer and does not get rid of this local minimum, making its final result not~as good as the results of our two-stage training. Moreover, it is  revealed that Stage III does not improve much of the performance of the two-stage training, which also in turn shows the effectiveness of the oracle-based two-stage training.

\subsection{Task-imbalanced Meta Learning}
\label{sec:metal}

In this section, we perform meta learning experiments in the few-short learning domain~\citep{ravi2017optimization}. 

\textbf{Experiment setting.} We follow the setting in MAML~\citep{finn2017model} for the few-shot learning tasks. 
 Each task is an N-way classification that contains meta-\{train, valid, test\} sets. On top of it, the macro dataset with multiple~tasks is split into train, valid and test sets. 
We consider the more realistic \textit{task-imbalanced} setting proposed by \citet{Na2020Learning}. Unlike the standard setting where the meta-train of each task contains $k$-shots for each class, here we vary the number of observation to perform $k_1$-
$k_2$-shot learning where $k_1 < k_2$ are the minimum/maximum number of observations per class, respectively. Build on top of MAML, we denote our variant as MAML-stop which learns how many adaptation gradient descent steps are needed for each task. 
 Intuitively, the tasks with less training data would prefer fewer steps of gradient-update to prevent overfitting. As we mainly focus on the effect of learning to stop, the neural architecture and other hyperparameters are largely the same as MAML. Please refer to Appendix~\ref{app:imba_maml} for more details. 

\noindent\textbf{Dataset.} We use the benchmark datasets Omniglot~\citep{lake2011one} and MiniImagenet~\citep{ravi2017optimization}. Omniglot consists of 20 instances of 1623 characters from 50 different alphabets, while MiniImagenet involves 64 training classes, 12
validation classes, and 24 test classes. We use exactly the same data split as~\citet{finn2017model}.~To construct the imbalanced tasks, we perform 20-way 1-5 shot classification on Omniglot and 5-way 1-10 shot classification on MiniImagenet. The number of observations per class in each meta-test set is 1 and 5 for Omniglot and MiniImagenet, respectively. For evaluation, we construct 600 tasks from the held-out test set for each setting. 
\begin{table}[h!]
\vspace{-4mm}
\centering
\caption{Task-imbalanced few-shot image classification. \label{tab:imba_maml}}
\begin{tabular}{ccc}
	\toprule
	& Omniglot & MiniImagenet  \\
	& 20-way, 1-5 shot & 5-way, 1-10 shot \\
\hline
MAML  & 97.96 $\pm$ 0.3\% & 57.20 $\pm$ 1.1\% \\
\hline
MAML-stop & $\mathbf{98.45 \pm 0.2\%}$ & $\mathbf{60.67 \pm 1.0\%}$ \\
\bottomrule
\end{tabular}	
\vspace{-4mm}
\end{table}

\begin{table*}[t]
\centering
\caption{Few-shot classification in vanilla meta learning setting~\citep{finn2017model} where all tasks have the same number of data points.}\label{meta-vanilla}
\resizebox{0.95\textwidth}{!}{%
\begin{tabular}{ccccccccc}
\toprule
& \multicolumn{2}{c}{Omniglot 5-way}  & \multicolumn{2}{c}{Omniglot 20-way} & \multicolumn{2}{c}{MiniImagenet 5-way} \\
\cmidrule(lr){2-3} \cmidrule(lr){4-5} \cmidrule(lr){6-7}
& 1-shot & 5-shot & 1-shot & 5-shot & 1-shot & 5-shot \\
\hline
MAML & 98.7 $\pm$ 0.4\% & 99.1 $\pm$ 0.1\% & 95.8 $\pm$ 0.3\% & 98.9 $\pm$ 0.2\%  & 48.70 $\pm$ 1.84\% & 63.11 $\pm$ 0.92\% \\
MAML-stop & {\bf 99.62 $\pm$ 0.22\%} & {\bf 99.68 $\pm$ 0.12\%} &  {\bf 96.05 $\pm$ 0.35\%} &  {\bf 98.94 $\pm$ 0.10 \%} &  {\bf 49.56 $\pm$ 0.82\%} &  {\bf 63.41 $\pm$ 0.80\%} \\
\bottomrule
\end{tabular}}
\vspace{-4mm}
\end{table*}

\noindent\textbf{Results.} Table~\ref{tab:imba_maml} summarizes the accuracy and the 95\%~confidence interval on the held-out tasks for each dataset. The maximum number of adaptation gradient descent steps is 10 for both MAML and MAML-stop. 
We can see the optimal stopping variant of MAML outperforms the vanilla MAML consistently. For a more difficult task on MiniImagenet where the imbalance issue is more severe, the accuracy improvement is 3.5\%. For completeness, we include the performance on vanilla meta learning setting where all tasks have the same number of observations in Table~\ref{meta-vanilla}. MAML-stop still achieves comparable or better performance. 

\subsection{Image Denoising}
In this section, we perform the image denoising experiments. More implementation details are provided in Appendix~\ref{app:image-denoising}.

\textbf{Dataset.} The models are trained on BSD500 (400 images) \citep{arbelaez2010contour}, validated on BSD12, and tested on BSD68 \citep{martin2001database}. We follow the standard setting in \citep{zhang2018dynamically,lefkimmiatis2018universal,zhang2017beyond} to add Gaussian noise to the images with a random noise level $\sigma\leqslant55$ during training and validation phases.

\textbf{Experiment setting.} We compare with two DL models, DnCNN \citep{zhang2017beyond} and UNLNet$_5$ \citep{lefkimmiatis2018universal},  and two traditional methods, BM3D \citep{dabov2007image} and WNNM \citep{gu2014weighted}. Since DnCNN is one of the most widely-used models for image denoising, we use it as our predictive model.
All deep  models including ours are considered in the \textit{blind} Gaussian denoising setting, which means the noise-level is not given to the model, while BM3D and WNNM require the noise-level to be known.

\vspace{-4mm}
\begin{table}[h!]
    \centering
    \caption{PSNA performance comparison. The sign * indicates that noise levels 65 and 75 do not appear in the training set.}
    \begin{tabular}{@{\hspace{1mm}}c@{\hspace{1mm}}|@{\hspace{1mm}}c@{\hspace{1mm}}|@{\hspace{1mm}}c@{\hspace{1mm}}|@{\hspace{1mm}}c@{\hspace{1mm}}||@{\hspace{1.5mm}}c@{\hspace{1.5mm}}|@{\hspace{1mm}}c@{}}
    \toprule
     $\sigma$ & {\small DnCNN-stop} & {\small DnCNN} & {\small UNLNet$_5$} & {\small BM3D} & {\small WNNM}   \\
    \midrule
    35 & \textbf{27.61} & 27.60 & 27.50 & 26.81 & 27.36 \\
    45 & \textbf{26.59} & 26.56 & 26.48 & 25.97 & 26.31 \\
    55 & \textbf{25.79} & 25.71 & 25.64 & 25.21 & 25.50 \\
    \hline \hline
    *65& \textbf{23.56} & 22.19 & - & 24.60 & \textbf{24.92} \\
    *75 & \textbf{18.62} & 17.90 & - & 24.08 & \textbf{24.39} \\
    \bottomrule
    \end{tabular}
    \label{tab:denoise_all_results}
    \vspace{-2mm}
\end{table}

\textbf{Results.} The performance is evaluated by the mean peak signal-to-noise ratio (PSNR). Table~\ref{tab:denoise_all_results} shows that DnCNN-stop performs better than the original DnCNN. Especially, for images with noise levels 65 and 75 which are \textbf{unseen} during training phase, DnCNN-stop {generalizes significantly better} than DnCNN alone. Since there is no released code for UNLNet$_5$, its~performances are copied from the paper~\citep{lefkimmiatis2018universal}, where results are not reported for $\sigma=65$ and $75$. For traditional methods BM3D and WNNM, the test is in the noise-specific setting. That is, the noise level is given to both BM3D and WNNM, so the comparison is not completely fair to learning based methods in blind denoising setting.

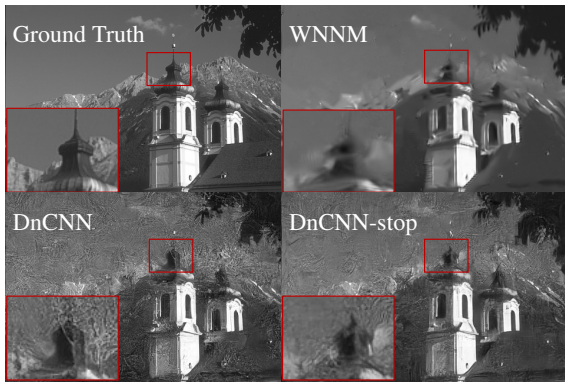
\begin{figure}[h!]
    \centering
    \vspace{2mm}
    \resizebox{0.9\linewidth}{!}{
    \begin{tabular}{@{}c@{}c@{}}
         \begin{picture}(115,74)
         \put(0,0){\includegraphics[width=0.5\linewidth]{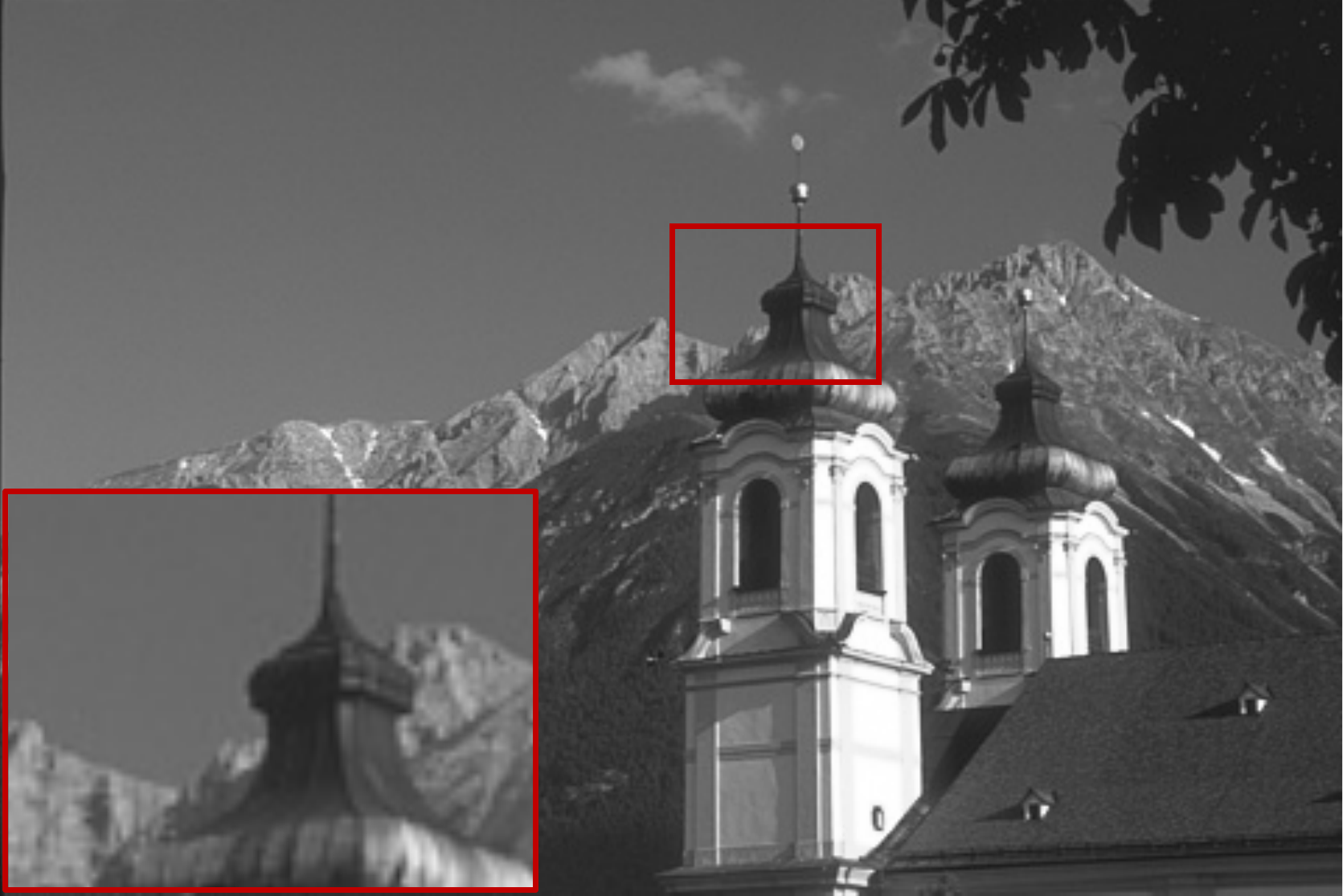}}
         \put(3,63){\color{white} Ground Truth}
         \end{picture}
         &  
         \begin{picture}(115,74)
         \put(0,0){\includegraphics[width=0.5\linewidth]{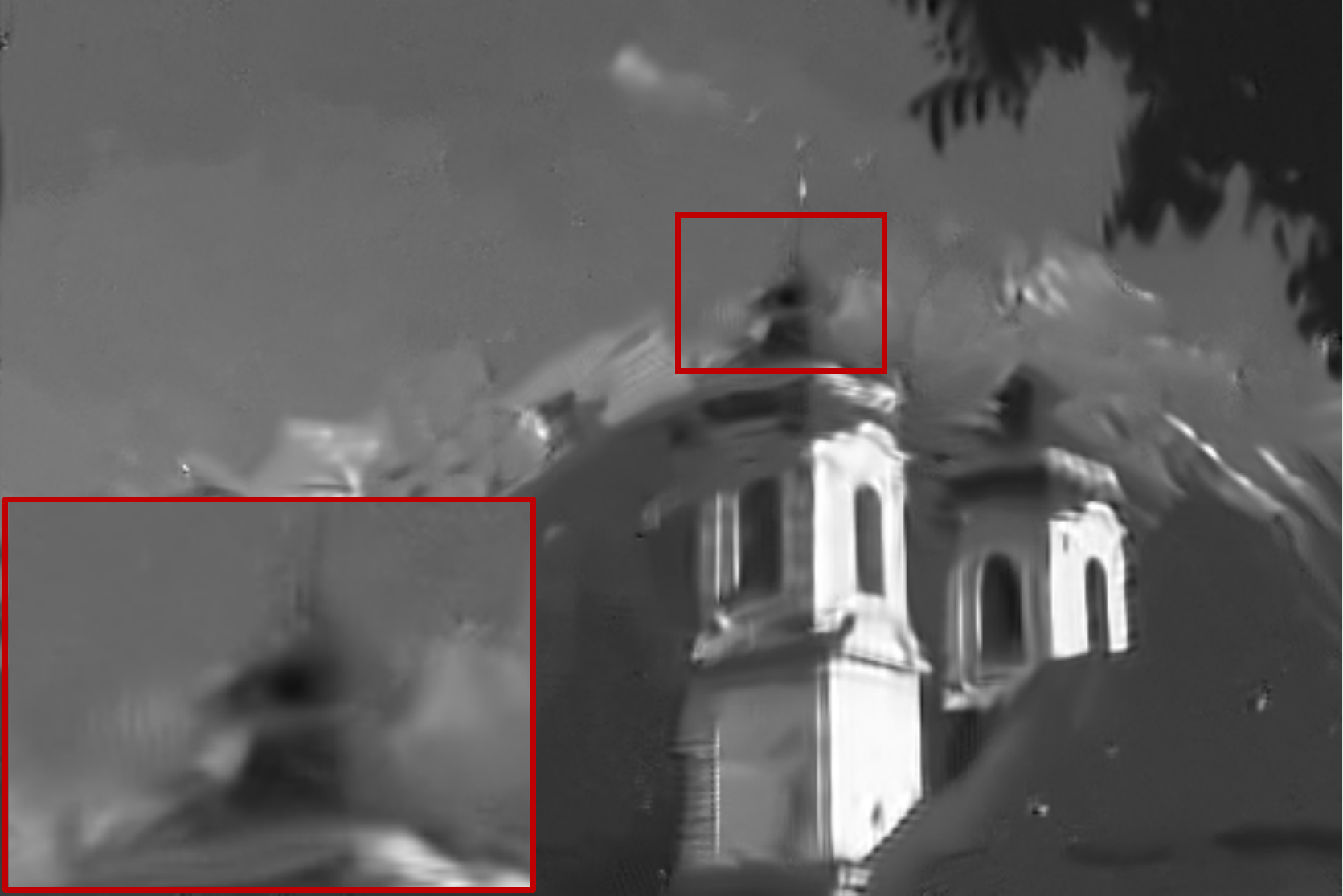}}
         \put(3,63){\color{white} WNNM}
         \end{picture}
         \\
         \begin{picture}(115,74.6)
         \put(0,0){\includegraphics[width=0.5\linewidth]{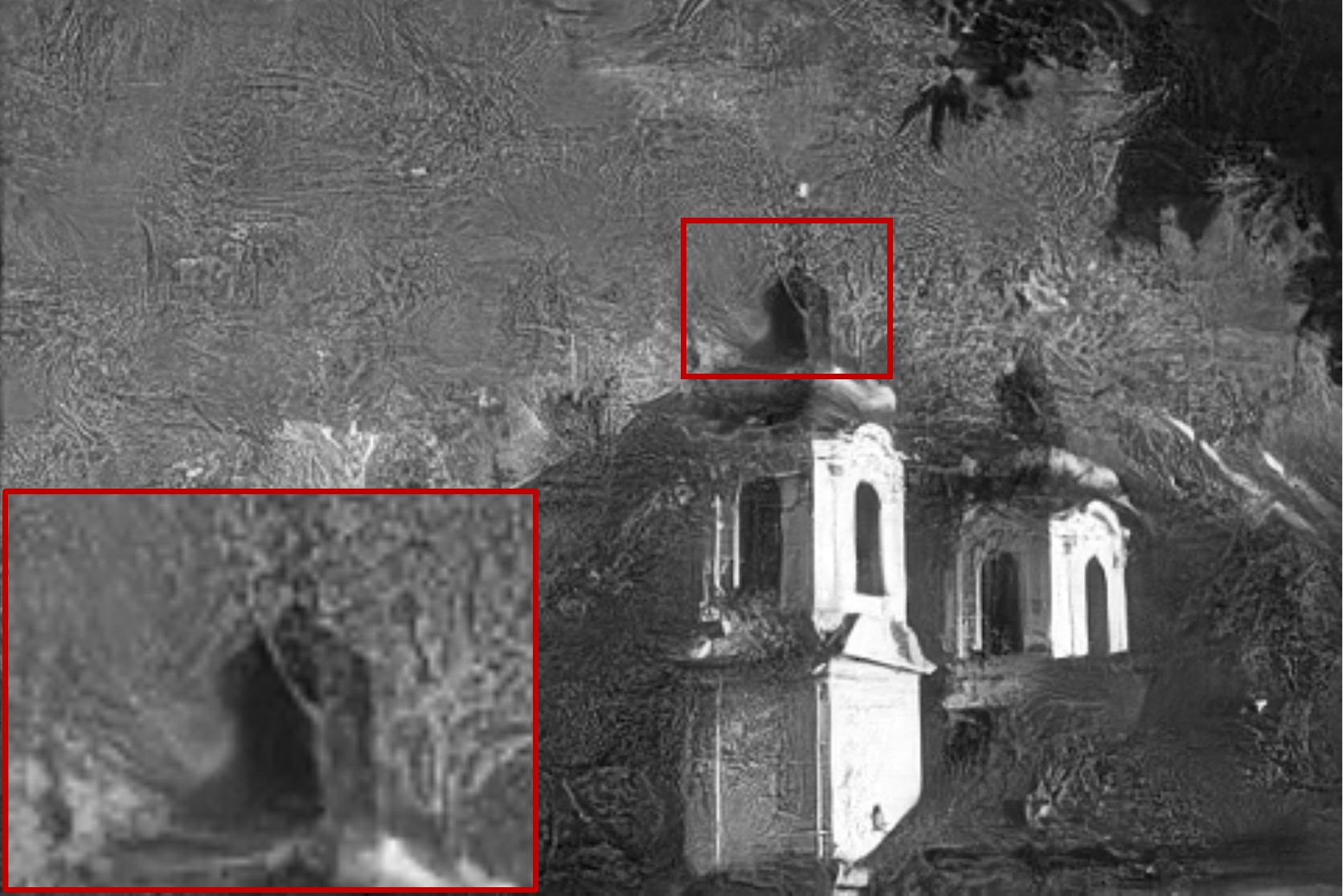}}
         \put(3,63){\color{white} DnCNN}
         \end{picture}
         &  
         \begin{picture}(115,74.6)
         \put(0,0){\includegraphics[width=0.5\linewidth]{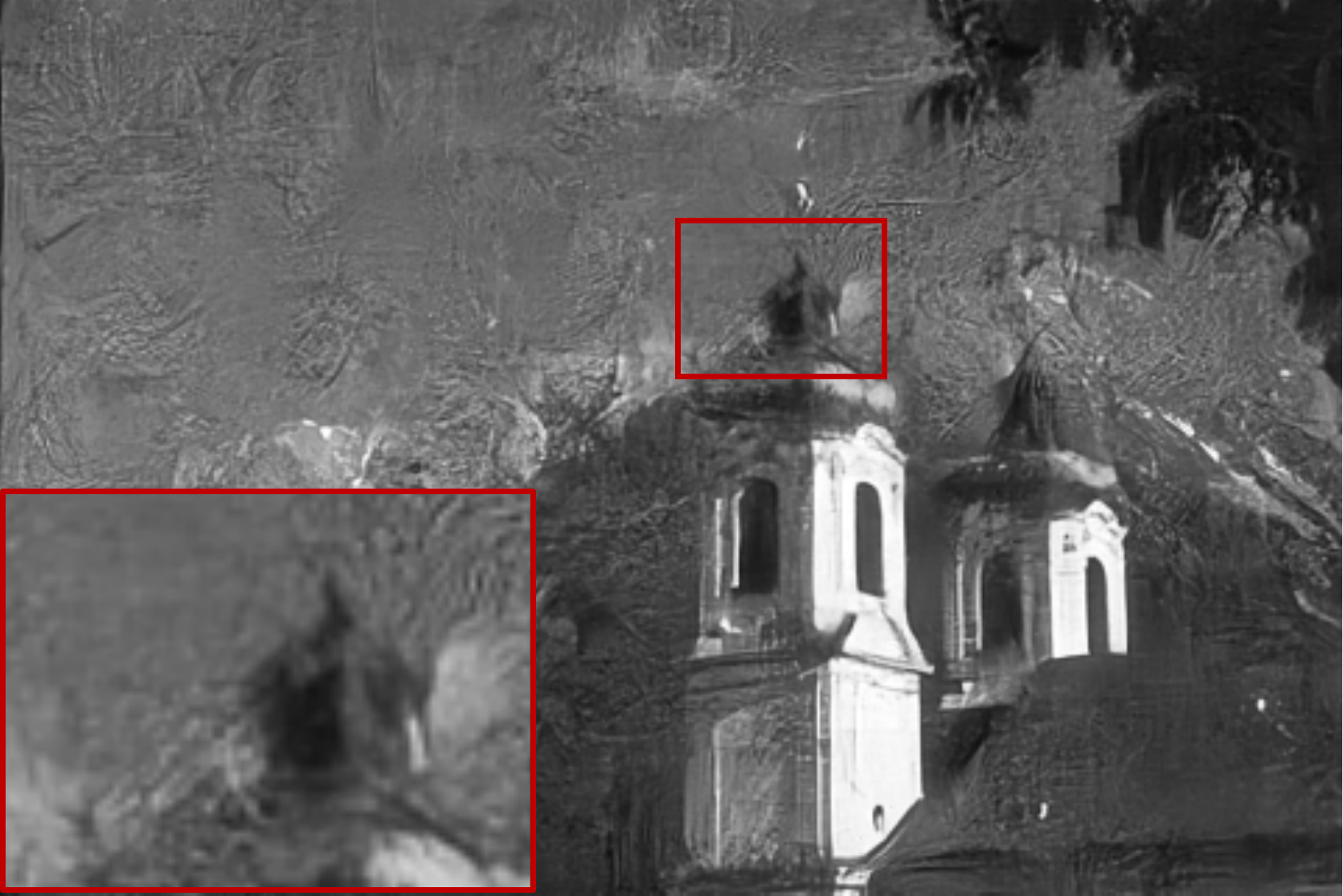}}
         \put(3,63){\color{white} DnCNN-stop}
         \end{picture}
         \\
    \end{tabular}}\vspace{-3mm}
    \caption{Denoising results of an image with noise level 65. (See Appendix~\ref{app:denoising-images} for more visualization results.)} \label{fig:my_label}
    \vspace{-3mm}
\end{figure}

\subsection{Image Recognition}
We explore the potential of our idea for improving the recognition performances on Tiny-ImageNet, using  VGG16 \citep{simonyan2014very} as the predictive model. With 14 internal classifiers, after Stage I training, if the oracle~$q_\theta^*$ is used to determine the stop time $t$, the accuracy of VGG16 can be improved to 83.26\%. Similar observation is provided in SDN \citep{kaya2019shallow}, but their loss $\sum_t w_t\ell_t$ depends on very careful hand-tuning on the weight $w_t$ for each layer, while we directly take an expectation using the oracle, which is more principled and leads to higher accuracy (Table~\ref{tab:image_rec}).
However, it reveals to be very hard to mimic the behavior of the  orcale $q_\theta^*$ by $\pi_\phi$ in Stage II, either due to the need of a better parametrization for $\pi_\phi$ or more sophisticated reasons. Our learned $\pi_\phi$ leads to similar accuracy as the heuristic policy~in SDN,
which becomes the bottleneck in our exploration. However, based on the large performance gap between the oracle and the original VGG16, our result still provides a potential direction for breaking the performance bottleneck of DL on image recognition.  
\vspace{-5mm}
\begin{table}[h!]
    \centering
    \caption{Image recognition with oracle stop distribution.}
    \begin{tabular}{c||c|c}
    \toprule
        VGG16 & SDN training & Our Stage I. training\\
    \hline
       58.60\% & 77.78\% (best layer) & 83.26\% (best layer)\\
    \bottomrule
    \end{tabular}
    \label{tab:image_rec}
    \vspace{-5mm}
\end{table}

\section{Conclusion}
In this paper,
we introduce a generic framework for modelling and training a deep learning model with input-specific depth, which is determined by a stopping policy $\pi_\phi$. Extensive experiments are conducted to demonstrate the effectiveness of both the model and the training algorithm, on a wide range of applications. In the future, it will be interesting to see whether other aspects of algorithms can be incorporated into deep learning models either to improve the performance or for better theoretical understandings.

\bibliographystyle{icml2020}

\appendix
\onecolumn
\section{Derivations}

\subsection{Proof of Lemma 1}
\label{app:lemma1}

\begin{proof}
Under the assumptions that 
\[
\ell(\vy,\vx_t;\theta):=-\log p_{\theta}(\vy|t,\vx);
\vspace{-1.5mm}
\]
and the prior $p(t|\vx)$ is a uniform distribution over $t$, the $\beta$-VAE objective can be written as
\begin{align*}
   \gJ_{\beta\text{-VAE}}&(\theta,q_\phi;\vx,\vy):= \nonumber\\
   &\E_{q_{\phi}}\log p_{\theta}(\vy|t,\vx) - \beta \text{KL}(q_{\phi}(t)||p(t|\vx)) \\
   =&-\E_{q_{\phi}}\ell(\vy,\vx_t;\theta)- \beta \E_{q_\phi(t)}\log\frac{q_{\phi}(t)}{p(t|\vx)}\\
   =&-\E_{q_{\phi}}\ell(\vy,\vx_t;\theta)- \beta \E_{q_\phi(t)}\log q_{\phi}(t) \\
   &+ \beta \E_{q_\phi(t)}\log {p(t|\vx)} \\
   =& -\rbr{\E_{q_{\phi}}\ell(\vy,\vx_t;\theta) - \beta H(q_\phi)}+\beta \E_{q_\phi(t)}\log \frac{1}{T}\\
   =& - \gL(\theta,q_\phi;\vx,\vy) - \beta \log T. 
\end{align*}
Since the second term $- \beta \log T$ is a constant, maximizing $\gJ_{\beta\text{-VAE}}(\theta,q_\phi;\vx,\vy)$ is equivalent to minimizing $\gL(\theta,q_\phi;\vx,\vy)$.
\end{proof}

\subsection{Equivalence of reverse KL and maximum-entropy RL}
\label{sec:kl-rl}
The variational distribution $q_\phi$ actually depends on the input instance $\vx$. For notation simplicity, we only write $q_\phi(t)$ instead of $q_\phi(t|\vx)$.
\allowdisplaybreaks
\begin{align}
    &\min_\phi \text{KL}(q_{\phi}(t)||q_{\theta}^*(t|\vy,\vx)) \\
    =&\min_\phi -\sum_{t=1}^{T}  q_{\phi}(t)  \log q_{\theta}^*(t|\vy,\vx)) - H(q_{\phi} ) \\
    =& \min_\phi  -\sum_{t=1}^{T}  q_{\phi}(t)  \log p_\theta(\vy|t,\vx)^{\beta})- H(q_{\phi} )\\
    &+  \sum_{t=1}^{T}  q_{\phi}(t)\log \sum_{\tau=1}^T p_{\theta}(\vy|\tau,\vx)^\beta \\
    =& \min_\phi  -\sum_{t=1}^{T}  q_{\phi}(t)  \log p_\theta(\vy|t,\vx)^{\beta}- H(q_{\phi} ) \\
    &+  \sum_{t=1}^{T}  q_{\phi}(t) C(\vx,\vy) \\
    =& \min_\phi  -\sum_{t=1}^{T}  q_{\phi}(t)  \log p_\theta(\vy|t,\vx)^{\beta}-H(q_{\phi} )\\
    &+ C(\vx,\vy) \\
    =& \min_\phi  -\sum_{t=1}^{T}  q_{\phi}(t)  \log p_\theta(\vy|t,\vx)^{\beta}- H(q_{\phi} ) \\
    =& \max_{\phi}\sum_{t=1}^{T}  q_{\phi}(t)  \log p_\theta(\vy|t,\vx)^{\beta} +H(q_{\phi} )\\
    =&  \max_{\phi}\sum_{t=1}^{T}  q_{\phi}(t) {\beta} \ell(\vy,\vx_t;\theta) +H(q_{\phi} ) \\
    =& \max_{\phi} \E_{t\sim q_{\phi}} \sbr{-\beta \ell(\vy,\vx_t;\theta) - \log q_{\phi}(t)}
\end{align}
Define the action as $a_t\sim\pi_t = \pi_\phi(\vx,\vx_t)$, the reward function as 
\begin{align*}
    r(\vx_t,a_t;\vy):= \begin{cases} -\beta \ell(\vy,\vx_t;\theta) & \text{if }a_t=1 \text{ (i.e. stop)},\\
    0 & \text{if }a_t=0 \text{ (i.e. continue)},
    \end{cases}
\end{align*}
and the transition probability as 
\begin{align*}
    P(\vx_{t+1}|\vx_t,a_t) = \begin{cases} 1 & \text{if }\vx_{t+1}=\gF_\theta(\vx_t)~\text{and}~a_t=0,\\
    0 & \text{else}.
    \end{cases}
\end{align*}
Then the above optimization can be written as
\begin{align}
 &\max_{\phi} \E_{t\sim q_{\phi}} \sbr{-\beta \ell(\vy,\vx_t;\theta) - \log q_{\phi}(t)} \\
  =  & \max_{\phi} \E_{\pi_\phi} \sum_{t=1}^T r(\vx_t,a_t;y) - \log  \pi_t(a_t|\vx,\vx_t)\\
    =& \max_{\phi} \E_{\pi_\phi}\sum_{t=1}^T \sbr{r(\vx_t,a_t;y) +H(\pi_t)}.
\end{align}

\section{Experiment Details}

\subsection{Learning To Learn: Sparse Recovery}
\label{app:lista}

\textbf{Synthetic data.} We follow \citet{chen2018theoretical} to choose $m = 250$, $n = 500$,  sample the entries of $A$ i.i.d. from the standard Gaussian distribution, i.e., $A_{ij} \sim \gN(0, \frac{1}{m})$, and then normalize its columns to have the unit
$\ell_2$ norm. To generate $\vy^*$, we decide each of its entry to be non-zero following the Bernoulli distribution with
$p_b = 0.1$. The values of the non-zero entries are sampled from the standard Gaussian distribution. The noise $\bm{\epsilon}$ is Gaussian white noise. The signal-to-noise ratio (SNR) for each sample is uniformly sampled from 20, 30 and 40. For the testing phase, a test set of 3000 samples are generated, where there are 1000 samples for each noise level. This test set is fixed for all experiments in our simulations.

\textbf{Evaluation metric.}  The performance is evaluated by NMSE (in dB), which is defined as $10\log_{10}(\frac{\sum_{i=1}^N
    \|\hat{\vx}^i-\vx^{*,i}\|_2^2}{\sum_{i=1}^N\|\vx^{*,i}\|_2^2})$ where $\hat{\vx}^i$ is the estimator returned by an algorithm or deep model.

\subsection{Task-imbalanced Meta Learning}
\label{app:imba_maml}

\subsubsection{Details of setup}

\paragraph{Hyperparameters} We train MAML with batch size 16 on Omniglot imbalanced and batch size 2 on MiniImagenet imbalanced datasets. In both scenario we train with 60000 of mini-batch updates for the outer-loop of MAML. We report the results with 5 inner SGD steps for Omniglot imbalanced and 10 inner SGD steps for MiniImagenet imbalanced with other best hyperparameters suggested in \citep{finn2017model}, respectively. For MAML-stop we run 10 inner SGD steps for both datasets, with the inner learning rate to be $0.1$ and $0.05$ for Omniglot and MiniImagenet, respectively. The outer learning rate for MAML-stop is $1e^{-4}$ as we use batch size 1 for training. 

When generating each meta-training dataset, we randomly select the number of observations within $k_1$ to $k_2$ for $k_1$-$k_2$-shot learning. The number of observations in test set is always kept the same within each round of experiment.

\subsubsection{Memory efficient implementation}

As our MAML-stop allows the automated decision of optimal stopping, it is preferable that the maximum number of SGD updates per each task is set to a larger number to fully utilize the capacity of the approach. This brings the challenge during training, as the loss on each meta-test set during training is required for \textit{each single} inner update step. That is to say, if we allow maximumly 10 steps of inner SGD update, then the memory cost for running CNN prediction on meta-test set is 10x larger than vanilla MAML. Thus a straightforward implementation will not give us a feasible training mechanism.

To make the training of MAML-stop feasible on a single GPU, we utilize the following techniques:
\begin{itemize}[leftmargin=*,nolistsep,nosep]
	\item We use stochastic EM for learning the predictive model, as well as the stopping policy. Specifically, we sample $t \sim q_{\theta}^*(\cdot|\vy,\vx)$ in each round of training, and only maximize $p_{\theta}(\vy|t,\vx)$ in this round. 
	\item As the auto differentiation in PyTorch is unable to distinguish between `no gradient' and `zero gradient', it causes extra storage for the unnecessary gradient computation. To overcome this, we first calculate $q_{\theta}^*(t|\vy,\vx)$ for each $t$ without any gradient storage (which corresponds to \texttt{no\_grad()} in PyTorch), then recompute $p_{\theta}(\vy|t,\vx)$ for the sampled $t$.  
\end{itemize}
With the above techniques, we can train MAML-stop almost as (memory) efficient as MAML. 

\subsubsection{Standard meta-learning tasks}

For completeness, we also include the MAML-stop in the standard setting of few-shot learning. We mainly compared with the vanilla MAML for the sake of ablation study.

\paragraph{Hyperparameters}

The hyperparameter setup mainly follows the vanilla MAML paper. For both MAML and MAML-stop, we use the same batch size, number of training epochs and the learning rate. For Omniglot 20-way experiments and MiniImagenet 5-way experiments, we tune the number of unrolling steps in $\cbr{5, 6, \ldots, 10}$, $\beta$ in $\cbr{0, 0.1, 0.01, 0.001}$ and the learning rate of inner update in $\cbr{0.1, 0.05}$. We simply use grid search with a random held-out set with 600 tasks to select the best model configuration.

\subsection{Image Denoising}
\label{app:image-denoising}

\subsubsection{Implementation Details}
When training the denoising models, the raw images were cropped and augmented into 403K $50*50$ patchs.  The training batch size was $256$. We used Adam optimizer with the initial learning rate as $1e-4$. We first trained the deep learning model with the unweighted loss for $50$ epochs. Then, we further train the model with the weighted loss for another $50$ epoches. After hyper-parameter searching, we set the exploration coefficient $\beta$ as 0.1. When training the policy network, we used the Adam optimizer with the learning rate as $1e-4$. We reused the above hyper-parameters during joint training. 

\subsubsection{Visualization}

\begin{figure}[h!]
    \centering
    \begin{tabular}{@{}c@{}c@{}c@{}}
         \includegraphics[width=0.33\linewidth]{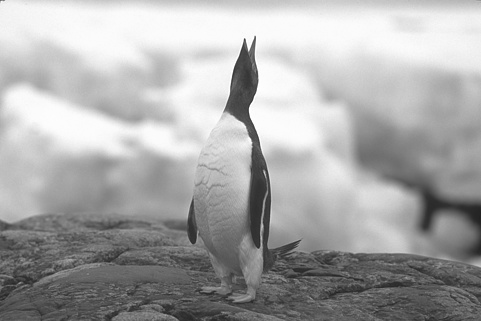}
         & 
         \includegraphics[width=0.33\linewidth]{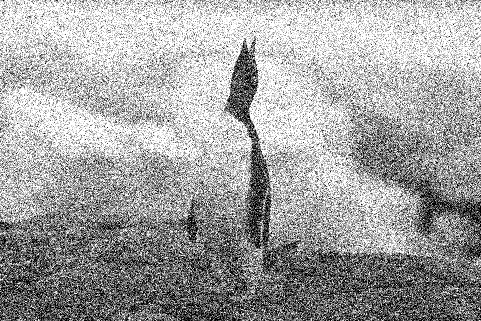}
        &
        \includegraphics[width=0.33\linewidth]{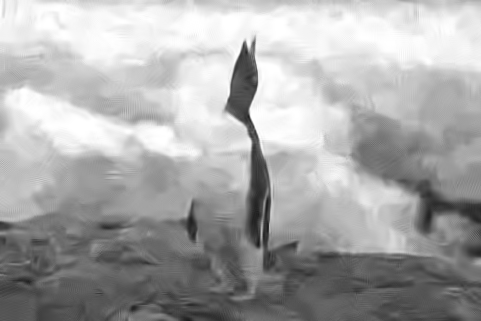}
         \\
         Ground Truth & Noisy Image & BM3D \\
         \includegraphics[width=0.33\linewidth]{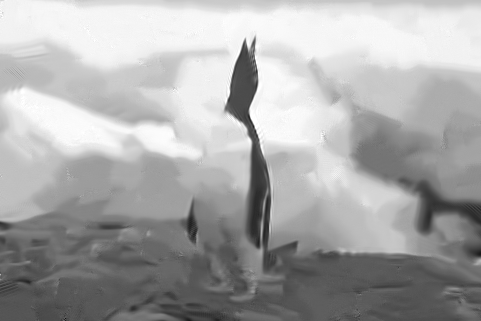}
         & 
         \includegraphics[width=0.33\linewidth]{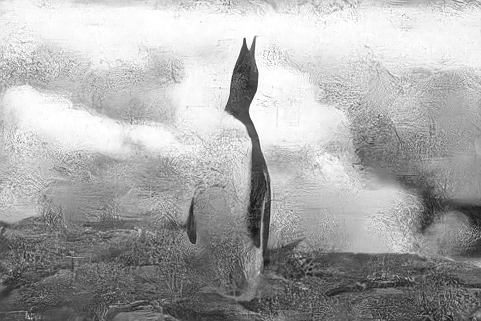}
        &
        \includegraphics[width=0.33\linewidth]{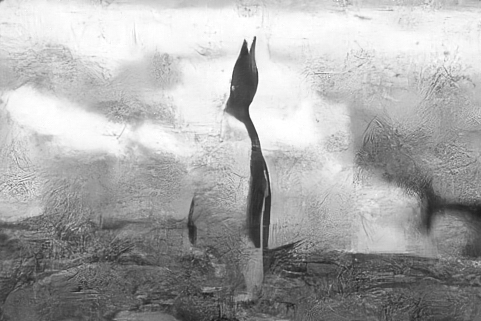}
         \\
         WNNM & DnCNN & DnCNN-stop
         \\
    \end{tabular}\vspace{-3mm}
    \caption{Denoising results of an image with noise level 65.}
\end{figure}

\begin{figure}[h!]
    \centering
    \begin{tabular}{@{}c@{}c@{}c@{}}
         \includegraphics[width=0.33\linewidth]{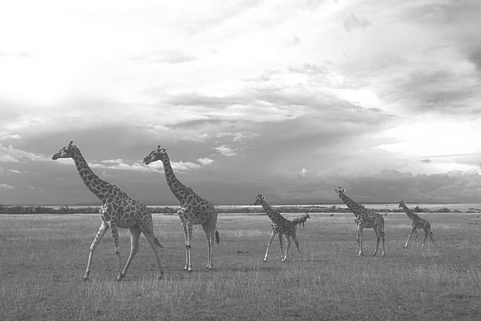}
         & 
         \includegraphics[width=0.33\linewidth]{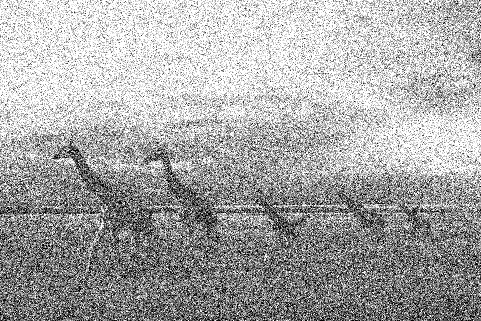}
        &
        \includegraphics[width=0.33\linewidth]{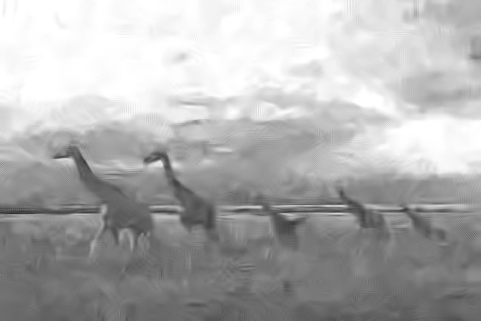}
         \\
         Ground Truth & Noisy Image & BM3D \\
         \includegraphics[width=0.33\linewidth]{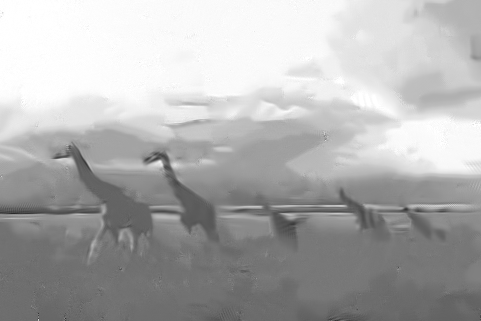}
         & 
         \includegraphics[width=0.33\linewidth]{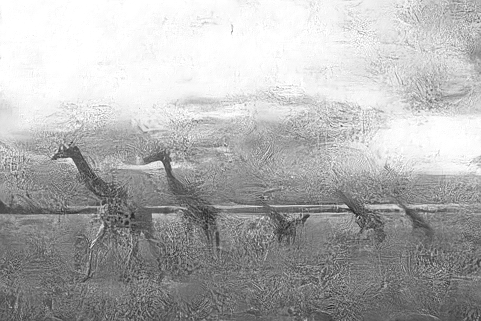}
        &
        \includegraphics[width=0.33\linewidth]{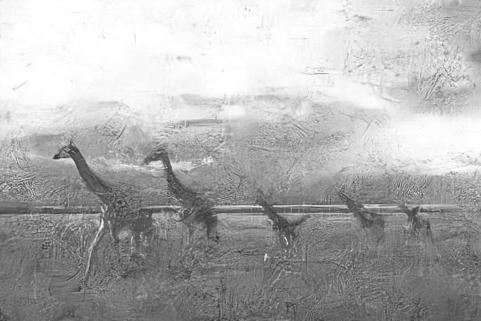}
         \\
         WNNM & DnCNN & DnCNN-stop
         \\
    \end{tabular}\vspace{-3mm}
    \caption{Denoising results of an image with noise level 65.}
\end{figure}

\label{app:denoising-images}

\subsection{Computing infrastructure}
Most of the experiments were run a hetergeneous GPU cluster. For each experiment, we typically used one or two V100 cards, with the typical CPU processor as Intel Xeon Platinum 8260L. We assigned 6 threads and 64 GB CPU memory for each V100 card to maximize the utilization of the card.

\end{document}